\documentclass[10pt,conference]{IEEEtran}
\IEEEoverridecommandlockouts

\usepackage{amsmath}
\usepackage{amssymb}
\usepackage{amsthm}
\usepackage{booktabs}
\usepackage{graphicx}
\usepackage{tikz}
\usepackage{algorithm}
\usepackage{algpseudocode}
\usepackage[hidelinks]{hyperref}
\usepackage{url}
\usetikzlibrary{arrows.meta,positioning}

\newtheorem{theorem}{Theorem}
\newtheorem{corollary}{Corollary}

\newcommand{\tool}{RobustPBE}

\begin{document}

\title{Fixed-Set Robustness in Programming by Example:\\Example Corruption and Semantic Partition Recovery}

\author{
\IEEEauthorblockN{Yuan Si and Jialu Zhang$^{*}$\thanks{Corresponding author: Jialu Zhang.}}\\
\IEEEauthorblockA{University of Waterloo, Waterloo, Canada\\
\texttt{yuan.si@uwaterloo.ca}, \texttt{jialu.zhang@uwaterloo.ca}}
}

\maketitle

\begin{abstract}
Programming-by-example systems infer programs from a small set of input-output examples.
Robust PBE work usually models wrong examples as samples from a stochastic noise process and then minimizes an expected or empirical loss.
This paper studies a different failure mode: an adversary who sees the synthesizer and chooses the examples whose corruption most damages the returned program.
We formalize fixed-set worst-case corruption for finite PBE version spaces, implement exact-within-bounded-pool and heuristic corruption searches for a string-transformation DSL, and introduce version-space partition aggregation (VPA), a defense that synthesizes on disjoint example groups and votes by semantic signatures.
The central claim is deliberately bounded and partly negative: low-margin PBE tasks have an adversarial robustness dimension that random-typo and noisy-PBE evaluations miss, while semantic partition aggregation helps only when the clean semantics keep a partition vote margin, which often fails on realistic tasks.
Evidence from curated/generated DSL tasks, accepted public SyGuS PBE\_SLIA slices, SYNTRA Playgol v2, and noisy-PBE objective baselines supports that boundary.
One curated edit flips all 8 spike tasks while 200-trial typo, DSL-pool, and distance-matched random controls succeed on 10.3\%, 11.0\%, and 16.7\%; generated margin-1 rows flip under budget 1 yet VPA recovers them; on public SyGuS the vote margin is near one, so an adaptive attacker drives VPA accuracy to zero; accepted public SyGuS slices move across exact-within-pool budget boundaries; and Playgol shows positive paired-bootstrap gaps against typo and same-pool random controls on the 141 accepted rows.
A small exact-output prompt harness over 20 controlled margin-1 tasks shows the same qualitative clean-to-attacked pattern across local and API models, but it is treated as a scope check, not a broad LLM benchmark.
\end{abstract}

\section{Introduction}

Programming by example (PBE) is attractive because it turns a few concrete input-output pairs into an executable transformation.
Spreadsheet users, data-cleaning tools, code assistants, and LLM prompts all rely on this interface.
The interface also creates a small and high-leverage attack surface.
When a PBE solver sees only three or five examples, a single corrupted example can change the set of programs that are consistent with the specification.
If the wrong program is simpler, better ranked, or better supported by the remaining examples, the synthesizer can return code that is exactly consistent with the corrupted examples and semantically wrong on future inputs.

Existing robust PBE work gives important tools for noisy data.
Handa and Rinard synthesize over noisy input-output examples with finite-tree-automata encodings and loss/complexity objectives, then formalize guarantees under noise-source assumptions~\cite{handa2020noisy,handa2021guarantees}.
Rose accelerates this objective with abstraction refinement~\cite{handa2021rose}.
RobustFill trains neural PBE models to tolerate realistic I/O noise such as typos~\cite{devlin2017robustfill}, and Raychev et al. learn programs from noisy datasets through a sampler and regularized generator~\cite{raychev2016noisy}.
These models are a natural first line of defense, yet they optimize for noise drawn from a source or observed in a dataset.
They leave open the fixed-set question: which small edit would an informed adversary choose after seeing the synthesizer?

This distinction matters in deployed workflows.
FlashFill's original paper opens from the observation that more than 500 million people use spreadsheets and describes an Excel add-in that synthesizes string transformations from input-output examples~\cite{gulwani2011flashfill}.
Microsoft describes PROSE as a framework for programming by examples that synthesizes a ranked set of DSL programs consistent with the examples~\cite{microsoft-prose}.
In a shared workbook or data-wrangling notebook, a user may correct a few rows, accept the inferred transformation, and apply it to downstream records without reading the hidden program.

Consider a shared sheet that normalizes payment records by example.
A co-editor or compromised account changes one output so the ranked program switches from preserving an account suffix to truncating it, or from one date interpretation to another.
The visible examples can still look locally plausible, while the wrong transformation propagates to unseen rows.
The incidents cited here are analogues for access, stakes, and silent propagation, not reports of PBE poisoning in the wild: Public Health England reported that 15,841 COVID-19 cases were delayed by a reporting pipeline issue~\cite{phe-delayed-cases}, and the U.S. Senate report on JPMorgan Chase's London Whale trades documents risk-model failures in a multi-billion-dollar loss~\cite{jpmorgan-whale-senate}.

The LLM setting adds a parallel few-shot surface.
Retrieval poisoning work shows that a single injected text can compromise a RAG system~\cite{corrupt-rag}, PoisonedRAG reports 90\% attack success with five malicious texts~\cite{poisonedrag}, and in-context poisoning shows that demonstrations can steer model behavior~\cite{he2025iclpoison}.
The PBE setting adds an exact semantic structure: the examples define a finite or enumerable version space of candidate programs, and the attack changes the decision boundary inside that space.

This paper studies the following question:
\emph{when does worst-case corruption of a fixed PBE example set create failures that random noisy-PBE models miss, and when can semantic aggregation recover?}
We answer within a deliberately explicit scope.
Our prototype implements a deterministic string DSL, exact-within-bounded-pool and heuristic corruption procedures, a semantic partition-aggregation defense, public SyGuS and Playgol evaluations, and a small exact-output LLM-PBE prompt harness.
The scope is string PBE plus exact-output LLM-PBE prompts.
FlashFill, PROSE, database-query, and code-generation benchmark families remain future evaluation targets.

\subsection{A Running Example}

Consider a small PBE task that extracts the first token from a name-like string:
\[
  \texttt{Ada Lovelace}\mapsto\texttt{Ada},\quad
  \texttt{Grace Hopper}\mapsto\texttt{Grace}.
\]
In a token DSL, the target is ``return token 1.''
A plausible rival is ``return token 2.''
If the second output is corrupted to \texttt{Hopper}, the target and the rival each satisfy one training example.
A deterministic tie-break can now choose the rival, and the synthesized program returns \texttt{Turing} on \texttt{Alan Turing}.
The corruption is small, local, and semantically plausible: the row still contains a token from the input.
It is also targeted.
A random typo in \texttt{Grace} is unlikely to make the second-token program win.

This example is intentionally simple.
The implemented attacks use larger candidate pools and held-out semantic scoring, but the same mechanism drives the experiments.
Strategic corruption searches for a replacement that moves mass from the target to a rival program.
Random-noise baselines perturb examples independently of this version-space objective.
VPA changes the unit of decision: it synthesizes on disjoint partitions and asks whether independently synthesized candidates still vote for the same semantics.

The core technical idea is to reason about margins in the PBE version space.
For a clean target program $p^\star$ and a rival $q$, only examples on which $p^\star$ and $q$ disagree can separate the two.
Changing enough of those outputs to match $q$ makes $q$ tie or beat the target under loss minimization.
This gives an attack recipe and also explains why many public examples are naturally robust: diverse examples increase the disagreement margin.
The defense, version-space partition aggregation (VPA), uses the same structure in the opposite direction.
It partitions examples into disjoint groups, synthesizes one candidate per group, evaluates those candidates on a finite vote domain, and selects the largest semantic-signature cluster.
The defense is conditional.
If the target semantics keep a vote margin, VPA is stable; if corruptions control a partition majority, VPA can fail.
The experiments include both outcomes.

The results support three claims.
First, worst-case corruption is a separate PBE robustness problem.
On a curated brittleness spike, one strategic output edit drops held-out accuracy from 1.000 to 0.000, while typo (0.103), same-pool (0.110), and distance-matched (0.167) random baselines rarely succeed.
On controlled generated tasks, all margin-1 tasks are flippable and margin-2 or margin-3 tasks resist budget-1 attacks.
Second, attack search is a real algorithmic issue.
For margin-2 and margin-3 tasks, bounded exact budget-2 search succeeds while greedy one-step search fails; beam search with width 32 matches bounded exact search in the measured sweep.
Third, semantic aggregation helps only in a narrow regime and is largely insufficient on realistic tasks.
VPA recovers all attacked generated margin-1 tasks and retains 0.922 mean attacked accuracy on Playgol, but on public SyGuS the clean vote margin is near one (certified radius near zero), so an adaptive attacker that knows the partitions and vote domain drives VPA accuracy to 0.000.

The paper makes four contributions.
It gives a worst-case corruption model for fixed PBE example sets, with attack success measured by held-out or bounded semantic accuracy.
It implements exact-within-bounded-pool and heuristic corruption search for output edits and finite searches for input edits, injection, and deletion.
It introduces VPA as a semantics-aware version-space analogue of partition aggregation, with a direct vote-stability certificate and explicit row-budget conditions.
It fixes seeds, resource guards, public benchmark provenance, generated paper tables, and a frozen record of known limitations for the reported evaluation.

\begin{figure}[t]
\centering
\begin{tikzpicture}[
  font=\scriptsize,
  box/.style={draw, rounded corners=1pt, align=center, inner sep=2pt, minimum height=0.68cm, text width=1.42cm},
  arr/.style={-{Stealth[length=1.5mm,width=1.1mm]}, line width=0.35pt}
]
\node[box] (clean) {$E^\star$\\clean\\examples};
\node[box, right=2.2mm of clean] (space) {$H$\\version\\space};
\node[box, right=2.2mm of space] (adv) {adversary\\budget $k$};
\node[box, right=2.2mm of adv] (corr) {$\hat E$\\worst-case\\corruption};
\node[box, right=2.2mm of corr] (synth) {$S(\hat E)$\\rival program;\\held-out\\error};
\node[box, below=6.5mm of clean] (def0) {$\hat E$\\defense\\input};
\node[box, right=2.2mm of def0] (groups) {disjoint\\groups};
\node[box, right=2.2mm of groups] (pergroup) {per-group\\synthesis};
\node[box, right=2.2mm of pergroup] (vote) {$\sigma_V$ vote\\over $V$};
\node[box, right=2.2mm of vote] (outcome) {recover\\or fail};
\draw[arr] (clean) -- (space);
\draw[arr] (space) -- (adv);
\draw[arr] (adv) -- (corr);
\draw[arr] (corr) -- (synth);
\draw[arr] (corr.south) -- ++(0,-2.2mm) -| (def0.north);
\draw[arr] (def0) -- (groups);
\draw[arr] (groups) -- (pergroup);
\draw[arr] (pergroup) -- (vote);
\draw[arr] (vote) -- (outcome);
\end{tikzpicture}
\caption{Threat model and VPA overview. The attack searches a budget-$k$ corruption that makes the synthesizer choose a semantically wrong rival. VPA changes the decision to a partition-level semantic vote and recovers only when the target signature keeps the largest cluster.}
\label{fig:overview}
\end{figure}
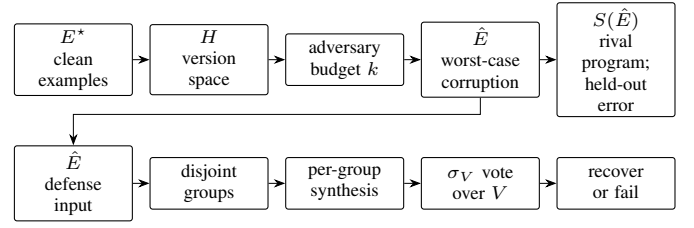

\section{Problem}

\subsection{PBE Under Fixed-Set Corruption}

Let $H$ be a DSL-defined hypothesis space and let $p^\star \in H$ be the intended program.
A clean example set is
\[
  E^\star = \{(x_i,p^\star(x_i))\}_{i=1}^n .
\]
A deterministic synthesizer returns
\[
  S(E) = \arg\min_{p\in H}(\ell_E(p), c(p), \tau(p)),
\]
where $\ell_E$ is training loss, $c$ is a simplicity cost, and $\tau$ is a fixed tie-break order.
The adversary has budget $k$ and produces $\hat{E}$ by changing at most $k$ examples.
The prototype implements output corruption, input corruption, example deletion, and example injection over bounded candidate sets.
Attack success is evaluated by semantic accuracy on held-out inputs or on a bounded finite semantic domain.

The adversary is white-box with respect to $H$, $S$, and the candidate-generation procedure.
This is the right model for public or easily imitated PBE systems, and it is also the conservative model for workflows where examples arrive from untrusted files or retrieved contexts.
The attacker leaves the final test inputs unchanged and changes the examples that induce the program.

Two modeling choices are worth making explicit.
First, the attack is measured against the intended program; the corrupted examples are the adversarial input to the synthesizer.
A successful attack can return a program that is perfectly consistent with the corrupted training set.
Second, the top-level budget counts example rows.
This matches common PBE interaction: a row, example, or demonstration is the unit a user or upstream source supplies.
Character-level distance is still used inside some candidate generators and noisy baselines, but the adversary's top-level budget counts examples touched.

\subsection{Why Stochastic Noise Is Insufficient}

Noisy-PBE objectives ask the solver to trade off data fit and complexity under an assumed, learned, or empirical noise model.
The same objective can be appropriate when user typos are independent or when noisy data are sampled from a stable process.
Fixed-set attacks have a different quantifier order:
\[
  \min_S \max_{\hat{E}: d(E^\star,\hat{E})\le k}
  \operatorname{err}(S(\hat{E}),p^\star).
\]
The maximization is over a small set of strategically chosen corruptions.
A single edit can be far more damaging than a typical typo because it targets a low-margin boundary between two programs.
Noisy-PBE baselines remain the most important comparison.
The prototype implements Handa--Rinard-style loss/complexity objectives over the same bounded finite version spaces.
The point is that expected-loss robustness and fixed-set worst-case robustness answer different questions.

\subsection{Margin Intuition}

For programs $p^\star$ and $q$, define the clean training disagreement set
\[
  \Delta_E(p^\star,q)=\{i: p^\star(x_i)\ne q(x_i)\}
\]
and its size $\mu_E(p^\star,q)$.
The synthesizer orders candidates lexicographically by $(\ell_E(p),c(p),\tau(p))$: training loss, simplicity cost, and deterministic tie-break key.
For a fixed rival $q$, define the pairwise targeted corruption threshold
\[
\begin{aligned}
  \kappa_E(p^\star,q)=\min\{k\le \mu_E :\;&
  (\mu_E-k,c(q),\tau(q)) \preceq{}\\
  & (k,c(p^\star),\tau(p^\star))\}.
\end{aligned}
\]
If the set is empty, the pair is not flippable by this targeted output construction.
The bounded-pool certificate used by the prototype is
\[
\begin{aligned}
  \kappa_E^{\mathcal C}(p^\star)=
  \min_{q\in \mathcal C,\;\mathrm{sem}(q)\ne \mathrm{sem}(p^\star)}
  \kappa_E(p^\star,q),
\end{aligned}
\]
followed by executing $S$ on the constructed corrupted set.

\begin{theorem}[Cost-aware targeted corruption threshold]
Fix $p^\star,q\in H$ and a deterministic solver order $(\ell,c,\tau)$.
If a budget-$k$ output attack changes $k$ examples in $\Delta_E(p^\star,q)$ to the outputs of $q$, then $p^\star$ has training loss $k$ and $q$ has training loss $\mu_E(p^\star,q)-k$ on the corrupted set.
The rival $q$ ties or beats $p^\star$ in the pairwise solver order exactly when $k\ge \kappa_E(p^\star,q)$.
With equal costs and no target-favoring tie-break, this reduces to crossing half the disagreement margin.
\end{theorem}

The certificate is intentionally bounded.
It identifies vulnerable rivals in the generated candidate pool and then validates the actual solver output.
Negative rows are exact only for that pool, which is why the paper reports candidate caps and phrases public negative rows as bounded-search results.
The public SyGuS results later show both sides of this boundary.
The 3/1 split has a small budget-1 vulnerability and a strong budget-2 vulnerability, while the 5/1 split has no budget-2 output vulnerability in the bounded candidate search and becomes vulnerable at budget 3.

The same margin view also gives a lower bound on what any deterministic solver can promise.
If two target programs disagree on the evaluation domain but their clean example sets differ on at most $2k$ labels, then there is a corrupted example set within budget $k$ of both worlds.
The solver receives the same input in both worlds and must return the same program, so it cannot be semantically correct in both.
This ambiguity bound is why the paper avoids an unconditional robustness claim.
For low-margin example sets, more examples or an explicit ambiguity warning is required.

\section{Method}

\subsection{Symbolic Synthesizer}

The implementation uses a deterministic FlashFill-style string DSL.
Atoms extract tokens, initials, prefixes, suffixes, constants, case transforms, and bounded structural templates.
Programs concatenate atoms.
The base DSL is deliberately small enough for bounded exact attack search on generated tasks.
Opt-in affix, numeric-span, and structured-template modes broaden public benchmark coverage.
For SyGuS PBE\_SLIA, the structured mode recognizes recurring name, phone, bike, and country-suffix schemas; it remains a bounded string-transformation DSL with explicit atoms and finite candidate sets.
This design keeps the core version-space behavior inspectable while supporting real public inputs.

The synthesizer enumerates candidates in a deterministic order and stores their training outputs, held-out outputs, cost, and semantic signatures where available.
This makes the attack implementation auditable.
For generated tasks, the target program and bounded semantic domain are known, so the prototype can check equivalence beyond the sampled held-out examples.
For public SyGuS tasks, the parser extracts examples from benchmark files and the structured DSL mines schema-specific atoms only from the training side.
For Playgol, the fixed corpus split is respected: three training examples induce the program and two test examples score it.
No generated table uses a row that was manually corrected after inspection.

\subsection{Attack Procedures}

The budget-1 output attack ranks DSL-realizable replacement outputs by the held-out damage of the programs that produce them.
The bounded exact output attack enumerates replacement combinations up to budget $k$ within a finite candidate set.
The finite candidate set is produced from source examples, mined constants, and DSL outputs.
For input edits, deletion, and injection, the prototype searches bounded candidate edits produced by the same DSL and corpus-specific generators.
The search is exact relative to those bounded candidate sets.
The paper reports candidate counts in the generated tables, separating bounded negative rows from claims about all possible strings.

The search loop has three stages.
It first constructs a replacement pool for each editable example.
It then synthesizes on each candidate corrupted set and records the returned program.
It finally evaluates the returned program on held-out or bounded semantic inputs.
For budget 1 this is a direct scan.
For budget $k>1$, bounded exact search enumerates combinations, while beam search keeps the most damaging partial corruptions after each depth.
The score used by the beam is held-out damage when the target is known and a conservative degradation proxy when only public held-out examples are available.

Algorithm~\ref{alg:strategic-output} is the output-corruption procedure used in the controlled and public output-attack rows.
The other implemented attack modes are finite input edits, example deletion, example injection, and paired prompt-example corruption for the LLM-PBE harness.
All exactness claims are relative to the generated candidate pool.

\begin{algorithm}[t]
\footnotesize
\caption{Strategic output-corruption attack}
\label{alg:strategic-output}
\begin{algorithmic}[1]
\Require Hypotheses $H$, synthesizer $S$, clean examples $E^\star$, budget $k$, beam width $w$
\Ensure Corrupted set $\widehat{E}$ maximizing held-out or bounded-semantic damage
\State Build replacement pool $C_i$ for each editable example $i$ from DSL outputs, constants, and source tokens
\State Initialize beam $B \gets \{(E^\star,\emptyset,0)\}$
\For{$d=1$ to $k$}
  \State $B' \gets \emptyset$
  \ForAll{$(E,J,s_{\mathrm{old}}) \in B$}
    \ForAll{$i \notin J$ and $y' \in C_i$}
      \State $\widehat{E} \gets E$ with output $i$ replaced by $y'$
      \State $p \gets S(\widehat{E})$; $s \gets$ damage of $p$ on held-out or bounded domain
      \State Add $(\widehat{E},J\cup\{i\},s)$ to $B'$
    \EndFor
  \EndFor
  \State $B \gets$ top $w$ elements of $B'$ by score
\EndFor
\State \Return highest-scoring $\widehat{E}$ in $B$
\end{algorithmic}
\end{algorithm}

Heuristic search matters because bounded exact output search grows quickly with budget and candidate cap.
The prototype implements one-step greedy and beam search.
The measured bounded exact-vs-heuristic sweep shows a non-monotone plateau: greedy misses all budget-2 attacks in the margin-2 and margin-3 settings, while beam search with width 32 finds the same attacks as bounded exact search.
This result is important for evaluation design.
If a paper reports only greedy attacks, it can falsely conclude that a task is robust.

\subsection{Version-Space Partition Aggregation}

VPA adapts partition aggregation from poisoning defenses to exact PBE semantics.
Given a corrupted example set $\hat{E}$, VPA partitions the examples into $m$ disjoint groups.
It runs the base synthesizer on each non-empty group, obtaining candidates $p_1,\ldots,p_m$.
For a finite vote domain $V$, each candidate gets a semantic signature
\[
  \sigma_V(p_j) = (p_j(x))_{x\in V}.
\]
VPA chooses the signature cluster with the largest number of votes and returns the simplest candidate in that cluster.
Algorithm~\ref{alg:vpa} gives the compact procedure used in the paper rows.

The VPA threat model is also white-box: for deterministic partitions, the attacker may know the groups and the vote domain.
The certificate is a vote-stability statement over signatures.
If the selected signature receives $v_{\top}$ votes and the largest runner-up receives $v_2$ votes, then changing at most
\[
  \rho = \left\lfloor\frac{v_{\top}-v_2-1}{2}\right\rfloor
\]
candidate signatures cannot change the selected signature.
Correctness also requires the target signature to be the selected clean signature and agreement on the vote domain to imply agreement on the evaluation domain.
The bridge from row budget to signature budget is explicit: when each row appears in one partition and synthesis on an untouched partition is unchanged, a budget-$k$ row attack can change at most $k$ partition signatures, so $k\le\rho$ is sufficient for signature stability.
The implementation enforces this bridge for the reported certificate checks by running synthesis on each partition's examples; the program list and schema vocabulary are fixed before partitioning, and examples outside a group are not used to score that group's candidate.
This bridge depends on partition-local synthesis over a hypothesis space whose constants and schema atoms are mined once from the clean specification before partitioning.
If a synthesizer re-mined constants or atoms from possibly corrupted examples at synthesis time, one corrupted row could alter several partitions' candidate pools, invalidating the one-row-to-one-signature bridge and its certificate; the prototype uses the fixed-vocabulary setting, which favors the defense because the attacker cannot inject new atoms through a corrupted output.
VPA is useful in recoverable regimes and fails when corruptions control the semantic vote.

The vote domain $V$ is chosen differently across settings.
Generated tasks use the bounded template domain.
Public SyGuS tasks use deterministic family domains when the structured recognizer can identify a schema; otherwise the held-out examples provide the measurable vote inputs.
Playgol uses the two supplied test examples for task-level scoring and the DSL-generated vote inputs for aggregation.
These choices are reported because the certificate is only as meaningful as the vote domain.
When $V$ is too small, two programs can share a signature while disagreeing elsewhere.
When $V$ captures the relevant schema, the signature is a useful semantic proxy.

\begin{algorithm}[t]
\footnotesize
\caption{Version-space Partition Aggregation}
\label{alg:vpa}
\begin{algorithmic}[1]
\Require Synthesizer $S$, corrupted examples $\hat E$, groups $G_1,\ldots,G_m$, vote domain $V$
\Ensure Aggregated program and vote-stability certificate
\State $A \gets \emptyset$
\For{$j=1$ to $m$}
  \If{$G_j$ is non-empty}
    \State $p_j \gets S(G_j)$; $\sigma_j \gets \sigma_V(p_j)=(p_j(x))_{x\in V}$
\State Add $(p_j,\sigma_j,c(p_j))$ to $A$
  \EndIf
\EndFor
\State Let $\sigma^\star$ be the most frequent signature in $A$
\State Let $v_{\top}$ be its vote count and $v_2$ the largest runner-up count
\State $\rho \gets \lfloor(v_{\top}-v_2-1)/2\rfloor$
\State \Return simplest $p_j$ with $\sigma_j=\sigma^\star$, together with $\rho$
\end{algorithmic}
\end{algorithm}

\begin{corollary}[Conditional semantic recovery]
If more partitions synthesize candidates with target signature $\sigma_V(p^\star)$ than any other signature, if at most $\rho$ partition signatures can be changed by the row-budget attack, and if agreement with $\sigma_V(p^\star)$ on $V$ implies agreement with $p^\star$ on the evaluation domain, then VPA returns a semantically correct program on that domain.
\end{corollary}

\section{Evaluation Design}

The evaluation answers four research questions.
RQ1 asks whether strategic example corruption behaves differently from random typo corruption.
RQ2 asks whether bounded exact search exposes attacks missed by greedy heuristics.
RQ3 asks whether VPA recovers in the regimes predicted by the vote-margin argument and fails when the precondition is broken.
RQ4 asks whether the pattern persists beyond curated DSL tasks, using public SyGuS, external Playgol, noisy-PBE baselines, and a scoped LLM-PBE prompt harness.

The research questions are ordered from mechanism to scope.
The first two use small controlled settings because they need ground truth about margins and bounded exact attacks.
The third question combines controlled positive cases and public negative cases because a defense paper must show where the defense stops working.
The fourth question asks whether the mechanism survives contact with public and external inputs.
This ordering keeps the story interpretable: controlled rows expose the mechanism, and public rows test whether the mechanism survives real inputs.

\subsection{Benchmarks}

The generated DSL benchmarks include random tasks and controlled-margin tasks.
The random tasks have diverse token patterns and usually high margins.
The controlled-margin tasks create a target and a rival separated by a chosen number of examples, making the theorem's threshold directly testable.
Each generated seed replicate uses fixed seeds and 30 tasks per scenario.

The curated spike is the cheapest Phase 0 test.
It exposes the mechanism before the broader measurements estimate its scope.
Its role is to show that the hypothesized mechanism exists before investing in the full prototype.
The generated random tasks then test the high-margin case: when examples are diverse, the same attack machinery often finds no damaging edit.
The controlled-margin tasks restore the low-margin condition in a reproducible way, which lets the paper attribute the effect to version-space geometry instead of to hand-picked strings.

The public SyGuS slice uses PBE\_SLIA string tasks from SyGuS-Comp.
The structured DSL currently solves all 79 measured tasks in the 80-file clean-coverage slice under the 3/1 split.
The 4/1 and 5/1 attack-split runs have 69 and 66 accepted measured tasks because some files lack enough usable examples.
The semantic check evaluates recognized name, bike, and phone families over bounded domains.

The public SyGuS rows use precise wording.
They measure the structured DSL on the accepted public slice.
They show that once a realistic public string benchmark is mapped into the implemented DSL, the same margin/budget behavior appears.
The evaluation records the SyGuS commit metadata, the parsed-file counts, and the accepted-task denominators.
It also runs a consistency audit so that mixed stale public inputs do not enter the paper tables.

The external Playgol measurement uses SYNTRA Playgol v2, 325 JSONL rows with explicit three-example training and two-example test splits.
The current token/affix/numeric-span DSL accepts 141 rows.
Rows outside the DSL are counted as coverage limits.

Playgol is important because it is independent of SyGuS and uses a different task source.
It is also a useful humility check.
The DSL covers less than half the corpus, and the strategic attack rate on accepted rows is small in absolute terms.
The scientific question is whether strategic corruptions beat random typos under the same accepted-task set.
The paired-bootstrap interval answers that question directly.

The LLM-PBE harness uses 20 controlled margin-1 tasks; each task has one clean and one attacked exact-output prompt, for 40 prompts per model.
The harness is a scope demonstration, not a general LLM benchmark: it tests whether the symbolic phenomenon transfers to exact-output prompting under a fixed prompt template and parser.
Each clean prompt and each attacked prompt is sent as an independent, context-isolated request; the clean and attacked variants of the same task never share a chat history or response context.
The reported answer metric gives credit to the first non-empty exact output and to deterministic explicit final answers such as an \texttt{Output:} line.
This parser is fixed before scoring and does not use a model judge.

\subsection{Baselines and Metrics}

Baselines include vanilla loss minimization, random typo corruption, matched random DSL-output replacement, edit-distance/row-budget matched random replacement, bounded exact attack, greedy attack, beam attack, VPA, leave-one-out voting, random-subset voting, trimmed loss, and Handa--Rinard-style noisy-PBE objectives.
The Handa--Rinard rows reproduce the published loss/complexity objectives over the prototype's bounded finite weighted version spaces, with state-compression and direct-enumeration parity tests.
They are objective-level reproductions because the public-implementation search found no public original implementation.
Random typo corruption is a natural-noise comparator.
The matched random controls reuse the same candidate pools and strategic edit-distance budgets without using the version-space damage ranking.
Together they separate natural-noise robustness from the stronger claim that the optimizer is exploiting version-space structure.
Claims about adversarial search strength rely on bounded candidate pools, exact or beam search, and candidate-count reporting.

Metrics are clean accuracy, attacked accuracy, attack success, VPA attacked accuracy or recovery, margin counts, and Wilson or bootstrap intervals where appropriate.
Attack success is the fraction of accepted tasks whose held-out or bounded semantic accuracy decreases under the attack.
For Playgol, paired bootstrap intervals compare strategic output attacks with random typo trials at the task level.
For public SyGuS proportions, Wilson intervals describe uncertainty over accepted tasks in the measured slice.

All reported rates are computed from persisted raw CSV or JSON summaries under \texttt{results/}.
The table-generation pipeline rebuilds every table from those persisted results.
The paper uses selected compact tables for readability, while the implementation keeps the full 28-table inventory.
This separation avoids two common failure modes: hiding negative rows in raw logs, and overwhelming the paper with every operational run.
The selected tables cover the curated and generated margin evidence, bounded exact-vs-heuristic search, public SyGuS budget boundaries, Playgol, representative baselines, and LLM-PBE prompts.
The omitted overview quantities remain in the text: the public SyGuS slice parses 80 files and measures 79 accepted 3/1 tasks, the 4/1 and 5/1 split runs measure 69 and 66 accepted tasks, Playgol parses 325 rows and accepts 141, and the LLM-PBE harness uses 40 prompts per model.

\section{Results}

\subsection{RQ1: Strategic Corruption Is Not Random Noise}

\paragraph{Curated Low-Margin Tasks.}
The first spike isolates the phenomenon.
The clean synthesizer succeeds on every task.
A single strategic output corruption succeeds on every task and drives held-out accuracy to 0.000.
In the matched-random rerun, typo corruption succeeds on 0.103 of trials, random DSL-output replacement from the same candidate pools succeeds on 0.110, and edit-distance/row-budget matched random replacement succeeds on 0.167.
This is the smallest experiment in the evaluation logs, and it gives the cleanest answer to RQ1: the strategic edit follows a different mechanism from the typo process.

The spike also guards against an easy alternative explanation.
If the attack were exploiting generic output damage, same-pool and distance-matched random replacements would often be comparable.
Instead, the strategic edit selects an output that is meaningful inside the DSL and that gives a specific rival program support.
The resulting error is semantic.
This is why held-out accuracy collapses even though the corrupted output may look like a valid token transformation.

\paragraph{Controlled Margins.}
The controlled generated benchmark tests whether the margin calculation predicts behavior beyond the spike.
Table~\ref{tab:generated-margin} shows the result.
All margin-1 tasks are flippable by one strategic edit, while typo, same-pool random, and distance-matched random controls succeed on 0.000, 0.007, and 0.065 of trials.
VPA recovers 1.000 attacked accuracy in this low-budget vote-margin regime.
Margin-2 and margin-3 tasks resist the budget-1 attack, matching the disagreement-threshold theorem.
Across five seeds, the margin-1 strategic success mean remains 1.000 with zero standard deviation; random train-5 tasks have strategic success mean 0.000.

The generated random row is equally important.
It shows that the attack machinery finds failures only when the version space contains a low-margin rival in the candidate pool.
Across the same five-seed replicate, random train-5 strategic success is 0.000; the table row is the displayed seed, where VPA attacked accuracy is 0.975.
This negative result narrows the claim.
Worst-case PBE corruption is a statement about low-margin regions of the version space and about evaluation protocols that must search those regions.

\begin{table}[t]
\centering
\caption{Curated and generated margin evidence for RQ1/RQ3. Typo, pool, and distance columns are random baseline success rates over 200 trials per task; attack accuracy is after the strategic edit.}
\label{tab:generated-margin}
\resizebox{\columnwidth}{!}{%
\begin{tabular}{lrrrrrrr}
\toprule
Setting & Tasks & Strat. & Typo & Pool & Dist. & Attack acc. & VPA acc. \\
\midrule
Curated spike & 8 & 1.000 & 0.103 & 0.110 & 0.167 & 0.000 & -- \\
Margin 1, train 5 & 30 & 1.000 & 0.000 & 0.007 & 0.065 & 0.000 & 1.000 \\
Margin 2, train 6 & 30 & 0.000 & 0.000 & 0.000 & 0.000 & 1.000 & 1.000 \\
Margin 3, train 7 & 30 & 0.000 & 0.000 & 0.000 & 0.000 & 1.000 & 1.000 \\
Random train 5 & 30 & 0.000 & 0.000 & 0.000 & 0.000 & 1.000 & -- \\
\bottomrule
\end{tabular}}
\end{table}

\subsection{RQ2: Bounded Exact Search Finds Attacks Greedy Search Misses}

Table~\ref{tab:exact-search} compares bounded exact, greedy, and beam search on controlled-margin tasks.
For margin 1 at budget 1, all search methods find the attack.
For margin 2 and margin 3 at budget 1, no attack exists in the bounded setting.
At budget 2, bounded exact search succeeds on both margin 2 and margin 3 rows, while greedy search succeeds on neither row.
Beam search with width 32 succeeds on both rows and matches bounded exact search on every measured row.

The practical lesson is that attack evaluation needs more than a one-step greedy adversary.
The failure mode is a plateau: the first edit can appear harmless until the second edit changes the loss ordering.
This is common in small PBE specifications because each example participates in several candidate explanations.
The implementation keeps bounded exact search for small cases and uses beam search as the scalable approximation.

This finding affects the interpretation of public negative rows.
A failed greedy attack is weak evidence.
A failed bounded exact or beam attack is stronger, but still tied to the candidate pool.
For this reason, the public tables report candidate counts and the text distinguishes bounded-search robustness from full semantic robustness.
That distinction is especially important for SyGuS strings, where the space of all possible replacement outputs is infinite.

\begin{table}[t]
\centering
\caption{Bounded exact-vs-heuristic output attack search on controlled margins.}
\label{tab:exact-search}
\resizebox{\columnwidth}{!}{%
\begin{tabular}{lrrrrrr}
\toprule
Margin & Budget & Tasks & Exact succ. & Greedy succ. & Beam succ. & Candidates \\
\midrule
1 & 1 & 30 & 1.000 & 1.000 & 1.000 & 15 \\
2 & 1 & 30 & 0.000 & 0.000 & 0.000 & 18 \\
2 & 2 & 30 & 1.000 & 0.000 & 1.000 & 153 \\
3 & 1 & 30 & 0.000 & 0.000 & 0.000 & 21 \\
3 & 2 & 30 & 1.000 & 0.000 & 1.000 & 210 \\
\bottomrule
\end{tabular}}
\end{table}

\subsection{RQ3: VPA Recovers Only with a Vote Margin}

The generated margin rows are the positive case.
The target semantics remain identifiable across partitions, and VPA returns the target signature.
The adaptive experiment in Table~\ref{tab:vpa-adaptive} replaces the attack score with direct damage to VPA's semantic output.
On generated margin-1 tasks, both vanilla-targeted and VPA-targeted attacks drive the vanilla synthesizer to 0.000 accuracy, yet VPA stays at 1.000 under known deterministic partitions and under eight hidden randomized partitions.
Because $\rho=0$ in these rows, this is empirical recovery under the measured adaptive search, not certified stability against one arbitrary signature change.
The shared $\rho=0$ radius does not imply identical empirical behavior: generated margin-1 uses budget 1 and the measured candidate pool does not realize a top-to-runner-up signature flip; SyGuS budget 2 lets the attacker control a partition majority.
On generated margin-2 tasks with budget 2, the same adaptive search confirms the sharp boundary: the attacker controls a partition majority, VPA accuracy is 0.000 with known partitions, and the hidden randomized protocol averages 0.125 with interval [0.125, 0.125].

The public SyGuS high-budget rows are the negative case.
With three singleton training partitions, a budget-2 output attack can corrupt a partition majority.
The VPA-targeted SyGuS run flips all 79 accepted tasks, VPA accuracy is 0.000 for both known and hidden partition protocols, and the signature-collision rate on the vote domain is 0.000.
The mean vote margin is only 1.03 and the median certificate radius $\rho$ is 0.
Randomizing singleton partitions does not create independent clean support; the VPA claim is limited to regimes where target semantics retain a partition vote margin.

\begin{table*}[t]
\centering
\caption{Adaptive VPA-targeted attack. Hidden rows average eight randomized partitions. Margin reports mean/p50 of $v_\top-v_2$; $\rho$ reports mean/p50. Cert. means $k\le\rho$ under the row-to-signature bridge; Emp. reports measured VPA recovery under the adaptive search.}
\label{tab:vpa-adaptive}
\resizebox{\textwidth}{!}{%
\begin{tabular}{llllrrrccrrr}
\toprule
Suite & Setting & Target & Part. & Tasks & Vanilla acc. & VPA acc. & Cert. & Emp. & Margin & $\rho$ & Coll. \\
\midrule
Generated & margin1, b1, g3 & vanilla & known & 30 & 0.000 & 1.000 & No & Yes & 1.00/1.00 & 0.00/0.00 & 0.000 \\
Generated & margin1, b1, g3 & VPA & known & 30 & 0.000 & 1.000 & No & Yes & 1.00/1.00 & 0.00/0.00 & 0.000 \\
Generated & margin1, b1, g3 & VPA & hidden & 30 & 0.000 & 1.000 & No & Yes & 1.00/1.00 & 0.00/0.00 & 0.000 \\
Generated & margin2, b2, g3 & VPA & known & 30 & 0.000 & 0.000 & No & No & 1.00/1.00 & 0.00/0.00 & 0.000 \\
Generated & margin2, b2, g3 & VPA & hidden & 30 & 0.000 & 0.125 & No & Partial & 1.00/1.00 & 0.00/0.00 & 0.000 \\
SyGuS & 3/1, b2, g3 & VPA & known & 79 & 0.000 & 0.000 & No & No & 1.03/1.00 & 0.01/0.00 & 0.000 \\
SyGuS & 3/1, b2, g3 & VPA & hidden & 79 & 0.000 & 0.000 & No & No & 1.03/1.00 & 0.01/0.00 & 0.000 \\
\bottomrule
\end{tabular}}
\end{table*}

Table~\ref{tab:sygus-budget} shows the broader public budget boundary.
With five training examples, the bounded output budget-2 attack flips none of the 66 accepted tasks, and VPA remains at 1.000; budget 3 flips all 66 and VPA again recovers none.
VPA accuracy is the held-out or bounded-semantic accuracy of the aggregated output after attack, while VPA recovery is the fraction of attacked tasks where the aggregate returns the clean target signature.
They coincide in these SyGuS rows because the accepted schema checks are binary for the reported split/task settings.

\begin{table*}[t]
\centering
\caption{Public SyGuS structured-DSL budget boundary. Rates are over accepted tasks in each row. The budget-1 output-attack row uses 100 ranked replacement outputs per example, and the budget-2/3 output-attack rows use 2 outputs per example, with full candidate-pool metadata in the evaluation logs.}
\label{tab:sygus-budget}
\resizebox{\textwidth}{!}{%
\begin{tabular}{llrrrrrrr}
\toprule
Split & Mode & Budget & Tasks & Attack succ. & Attack acc. & VPA acc. & VPA rec. & Wilson 95\% CI for attack succ. \\
\midrule
3/1 & output & 1 & 79 & 0.013 & 0.987 & 0.987 & 0.987 & [0.002, 0.068] \\
3/1 & input & 2 & 79 & 0.266 & 0.734 & -- & -- & [0.181, 0.372] \\
3/1 & input & 3 & 79 & 0.873 & 0.127 & 0.127 & 0.127 & [0.782, 0.930] \\
3/1 & output & 2 & 79 & 1.000 & 0.000 & 0.000 & 0.000 & [0.954, 1.000] \\
3/1 & delete & 2 & 79 & 0.013 & 0.987 & -- & -- & [0.002, 0.068] \\
3/1 & inject & 2 & 79 & 0.013 & 0.987 & -- & -- & [0.002, 0.068] \\
5/1 & output & 2 & 66 & 0.000 & 1.000 & 1.000 & 1.000 & [0.000, 0.055] \\
5/1 & output & 3 & 66 & 1.000 & 0.000 & 0.000 & 0.000 & [0.945, 1.000] \\
\bottomrule
\end{tabular}}
\end{table*}

\subsection{RQ4: Public and External Corpora}

\paragraph{Public SyGuS.}
On the accepted structured SyGuS slice, budget boundaries appear sharply, while frozen held-out coverage remains low.
The current measured 3/1 slice accepts 79 of 79 tasks after the targeted country-suffix atom, and the bounded-family semantic check confirms the clean row on recognized domains.
Budget-1 output attacks have a small but real counterexample: 1 of 79 accepted 3/1 tasks.
Moving to 4/1 and 5/1 splits removes that budget-1 attack in the bounded search, with 0 of 69 and 0 of 66 successes.
The same benchmark then becomes vulnerable again at higher budgets, matching the expected margin boundary.

The single budget-1 public vulnerability is useful despite its small rate.
It arose from a country-suffix pattern outside the earlier clean DSL coverage.
The targeted atom improved clean coverage and exposed a sharper boundary: the clean system became stronger, yet the attack could still identify a low-margin row under the 3/1 split.
A synthesizer's clean coverage and attack surface are coupled; robustness must be re-evaluated whenever the hypothesis space changes.

The frozen-DSL held-out check separates this atom-selection step from later evaluation.
Using the same 379,910-program DSL and only the development constants, a sorted held-out SyGuS slice at files 81--110 accepts 6 of 30 parsed tasks, for 0.200 clean coverage.
On those accepted rows, budget-2 output attack success is 1.000, attacked accuracy is 0.000, and VPA attacked accuracy is 0.000.
This row supports the budget-boundary mechanism on accepted held-out tasks, while the 24 clean-loss rejections make DSL coverage a visible limitation.

\paragraph{Playgol.}
Playgol is the main non-SyGuS external check.
The DSL cleanly covers 141 of 325 rows, a 0.434 accepted-row rate with Wilson 95\% interval [0.381, 0.488].
On those accepted rows, strategic output attack success is 8/141, or 0.057 with Wilson interval [0.029, 0.108].
Random typo success is 7/705, or 0.010 with interval [0.005, 0.020].
Same-pool random replacement reuses the strategic candidate pools without the damage ranking and succeeds on 5/705 trials, or 0.007 with interval [0.003, 0.016].
The paired-bootstrap estimate is positive for strategic minus typo success, 0.047 with 95\% interval [0.017, 0.082], and for strategic minus same-pool success, 0.050 with interval [0.018, 0.088].
VPA exact recovery is 129/141 and the task-bootstrap mean attacked accuracy is 0.922 with interval [0.876, 0.961].
These are modest rates.
They support RQ4 because the strategic-vs-random-control gaps are positive on an external corpus and VPA preserves high attacked accuracy on the accepted subset.

The coverage denominator is part of the result.
Reporting attack success on the 141 accepted rows alone would imply corpus-wide coverage.
The paper reports both coverage and attack evidence.
The coverage audit records 180 clean-loss rejections and 4 clean-generalization rejections; examples in the rejected slice include transformations requiring substring deletion, lookup, or character-level operations outside the current DSL.
This makes the current scope clear: Playgol supports external validity for the accepted token/affix/numeric-span subset, while the 184 rejected rows motivate richer DSL support after submission.

\begin{table}[t]
\centering
\caption{External Playgol evidence.}
\label{tab:playgol}
\resizebox{\columnwidth}{!}{%
\begin{tabular}{lrrr}
\toprule
Measure & Count & Estimate & 95\% interval \\
\midrule
Clean coverage accepted & 141/325 & 0.434 & [0.381, 0.488] \\
Strategic output success & 8/141 & 0.057 & [0.029, 0.108] \\
Random typo success & 7/705 & 0.010 & [0.005, 0.020] \\
Same-pool random success & 5/705 & 0.007 & [0.003, 0.016] \\
Strategic minus typo & -- & 0.047 & [0.017, 0.082] \\
Strategic minus same-pool & -- & 0.050 & [0.018, 0.088] \\
VPA exact recovery & 129/141 & 0.915 & [0.857, 0.951] \\
VPA attacked mean acc. & -- & 0.922 & [0.876, 0.961] \\
\bottomrule
\end{tabular}}
\end{table}

\subsection{Noisy-PBE Baselines}

The baseline rows help separate three effects.
On the controlled target-favored regime, Handa--Rinard 0/1-bound and trimmed-loss objectives can preserve the target under the attack because their objective pays for ignoring the corrupted point.
On the rival-favored regime, the same family of methods can fail because the objective and tie-break favor the rival.
On the public structured 3/1 budget-1 row, vanilla synthesis already has attacked accuracy 0.987 because the benchmark margin is high; Handa--Rinard 0/1-bound, trimmed loss, and VPA also report 0.987.
This public row is a high-margin consistency check.
The weaker substitution and voting baselines expose underidentification under very small training splits.

The comparison is objective-level.
The prototype implements the published loss/complexity objectives over bounded finite weighted version spaces and tests parity against direct enumeration.
The public implementation search found no original Handa/Rinard implementation, so the comparison targets the published objectives directly.
This is adequate for the min-max claim in this paper: the relevant question is how published noisy-data objectives behave under fixed-set worst-case corruptions in the same version space.
The edit-distance row is lower on the public 3/1 setting because a small string edit can still move support to a simpler rival in this DSL, while the 0/1-bound and trimmed objectives ignore or cap that point under the bounded candidate pool.
The target-favored and rival-favored rows are generated from the same finite version space with the loss/cost order changed to favor the target or the rival, so they are mechanism checks, not prevalence estimates.

The baseline table also prevents overclaiming VPA.
In the public 3/1 budget-1 setting, VPA, vanilla, trimmed loss, and Handa--Rinard 0/1-bound have the same 0.987 attacked accuracy.
The bounded attack found one vulnerability, so several methods share the same high attacked accuracy.
The controlled target-favored row is where VPA and robust objectives show recovery, and the rival-favored row is where the same approaches can fail.
Together these rows keep the conclusion tied to the data.

\begin{table}[t]
\centering
\caption{Representative baseline outcomes. Controlled rows summarize attacked accuracy in the target-favored and rival-favored constructions; public rows are the structured SyGuS 3/1 budget-1 attacked setting.}
\label{tab:baselines}
\resizebox{\columnwidth}{!}{%
\begin{tabular}{lrrr}
\toprule
Method & Target-fav. & Rival-fav. & Public 3/1 \\
\midrule
Vanilla & 0.000 & 0.000 & 0.987 \\
Trimmed loss & 1.000 & 0.000 & 0.987 \\
H-R 0/1 bound & 1.000 & 0.000 & 0.987 \\
H-R edit distance & 0.000 & 0.000 & 0.949 \\
Leave-one-out vote & 0.000 & 0.000 & 0.127 \\
Subset vote & -- & -- & 0.139 \\
VPA & 1.000 & 0.000 & 0.987 \\
\bottomrule
\end{tabular}}
\end{table}

\subsection{LLM-PBE Prompt Results}

The LLM-PBE rows are a scope demonstration.
They test whether exact-output few-shot prompting shows the same strategic-example sensitivity under the controlled margin-1 prompt set.
The local rows use Ollama models and checked-in Modelfiles where needed.
The API rows use \texttt{Qwen3.7-plus}, \texttt{Kimi-k2.7-code}, \texttt{GLM-5.2}, \texttt{gpt-5.5}, \texttt{claude-opus-4-8}, and \texttt{gemini-3.1-pro-high}, with a 512-token cap.
The API collection uses temperature 0 and one stateless request per prompt.
All six API models degrade sharply: Qwen3.7-plus falls from 0.750 to 0.150, Kimi-k2.7-code from 1.000 to 0.000, GLM-5.2 from 0.900 to 0.000, and \texttt{gpt-5.5}, \texttt{claude-opus-4-8}, and \texttt{gemini-3.1-pro-high} each from 1.000 to 0.000.
The local Qwen 14B and CPU-only Qwen 32B rows also fall from 1.000 to 0.000, and the reproducible DeepSeek 16B 8k row falls from 0.800 to 0.050.

These rows support only a prompt-harness claim: PBE examples in a prompt can form the same kind of small, high-leverage decision surface.
They are not used as prevalence estimates for LLM-PBE robustness.

The API rows add stronger model coverage and use the same 40-prompt controlled task set as the local rows.
With the task distribution fixed, the clean-to-attacked gap is attributable to example corruption.
Answer extraction is deterministic and conservative: verbose responses receive credit only when a clear final answer can be parsed without a learned judge.

\begin{table}[t]
\centering
\caption{LLM-PBE clean-to-attacked answer accuracy. Each model has 20 tasks, with one clean and one attacked prompt per task, for 40 prompts total.}
\label{tab:llm}
\resizebox{\columnwidth}{!}{%
\begin{tabular}{lrr}
\toprule
Model & Clean answer & Attacked answer \\
\midrule
Qwen2.5-Coder 1.5B & 0.500 & 0.250 \\
Qwen2.5-Coder 7B & 0.950 & 0.100 \\
Qwen2.5-Coder 14B & 1.000 & 0.000 \\
Qwen2.5-Coder 32B CPU & 1.000 & 0.000 \\
DeepSeek-Coder-V2 16B & 0.650 & 0.350 \\
DeepSeek-Coder-V2 16B 8k & 0.800 & 0.050 \\
Qwen3.7-plus API & 0.750 & 0.150 \\
Kimi-k2.7-code API & 1.000 & 0.000 \\
GLM-5.2 API & 0.900 & 0.000 \\
\texttt{gpt-5.5} API & 1.000 & 0.000 \\
\texttt{claude-opus-4-8} API & 1.000 & 0.000 \\
\texttt{gemini-3.1-pro-high} API & 1.000 & 0.000 \\
\bottomrule
\end{tabular}}
\end{table}

\section{Failure Anatomy}

The public SyGuS results are sharp because both the hypothesis class and the attack budget matter.
The single budget-1 3/1 vulnerability is task \texttt{11440431}, a country-suffix transformation whose clean structured program is \texttt{strip\_country\_suffix}.
With the widened candidate pool, one output corruption makes the solver return \texttt{tok0}; held-out accuracy is 0.000, and the bounded country-suffix semantic check over 18 points gives semantic accuracy 0.333.
This is not a DSL-success story alone: adding the atom improves clean coverage while preserving a low-margin boundary, and singleton VPA partitions can still underidentify the target signature.

The same mechanism explains the public budget boundary.
Under the 3/1 split, budget 1 flips 1 of 79 accepted tasks and budget 2 flips all 79; under the 5/1 split, budget 2 flips none of 66 accepted tasks and budget 3 flips all 66.
More independent examples raise the target margin, while coordinated output edits can cross the majority threshold.
Candidate-pool size is also part of the claim: a 20-candidate pool misses the country-suffix vulnerability, while the 100-candidate pool finds it.
The evaluation reports candidate counts, split metadata, and bounded-search negatives so that ``no attack found'' is never read as unbounded robustness.

\section{Discussion}

The main outcome is a boundary, not a prevalence claim.
Low-margin PBE tasks are brittle; diverse public examples often raise the margin; bounded exact or beam attacks are needed when a semantic flip requires coordinated edits.
VPA recovers when partitions contain enough independent clean support and fails when the attacker controls the semantic vote, matching the cost-aware margin and vote-stability conditions.
The negative rows are part of the claim: they show when the bounded adversary is ineffective, when higher budgets reopen the boundary, and when VPA should report under-specification instead of returning an overconfident program.

The adaptive VPA rows make that boundary sharper.
They rule out the optimistic reading that the earlier attacks recovered only because they targeted the vanilla synthesizer.
In generated margin-1 tasks, an attacker that scores candidate corruptions by VPA damage still leaves the semantic vote intact.
In generated margin-2 and public SyGuS budget-2 rows, the same white-box search drives VPA to the wrong semantic cluster once the row budget can control a partition majority.
Hidden randomized partitions help only when the random grouping creates enough clean support; with three singleton groups, randomization mostly permutes the same underidentified votes.
For deployment, this means VPA is best treated as a certificate-producing aggregation rule.
It can justify a returned transformation when vote signatures agree with margin, and it can surface an ambiguous specification when the margin disappears.

For PBE tools, report the split, corruption budget, candidate pool, semantic denominator, exact prompts, parser, model family, and clean-to-attacked gap; these quantities decide whether negative rows or prompt results are meaningful.

\section{Implementation Notes}

The evaluation fixes random seeds, public benchmark provenance, resource guards, and validation checks before table generation. Resource guards cover public SyGuS sweeps and local LLM runs; low-memory checks are used for the paper tables, while heavier runs are treated as scope-limited experiments.

\section{Threats to Validity}

\paragraph{DSL Coverage.}
The bounded string DSL solves all 79 measured public SyGuS tasks in the current 3/1 slice and accepts 141 of 325 Playgol rows, while the frozen-dev-constant held-out slice accepts 6 of 30 parsed tasks.
Claims are limited to accepted DSL-covered rows and the prompt harness; FlashFill/PROSE, NosDAQ-style query tasks, and CodeARC/MBPP-style code-generation suites remain future work.

\paragraph{Attack Candidate Bounds.}
Exact attacks are exact within bounded candidate sets; generated tasks are close to the intended finite setting, while public strings remain bounded-search claims.
Negative attack rows should be read with their candidate caps and train splits, including the public 5/1 budget-2 output row.

\paragraph{VPA Adaptivity.}
The VPA certificate covers white-box deterministic partitions only when row budget maps to a bounded number of changed partition signatures; adaptive experiments maximize VPA failure, add hidden randomized partitions, preserve generated margin-1 recovery, and confirm failure when the budget controls a partition majority.
The defense claim is limited to vote-margin regimes with enough independent clean support.

\paragraph{Random Baselines.}
Random typo rows model natural stochastic noise; matched DSL-output and edit-distance controls reuse attack pools and edit budgets, remain far below strategic success in curated and generated margin-1 rows, and remain finite-pool controls.
A broader semantic random-output process is outside the current evaluation.

\paragraph{Semantic Equivalence.}
Generated semantic checks evaluate bounded finite template domains; public SyGuS checks evaluate recognized family schemas over deterministic name, bike, and phone domains.
They are stronger than one held-out example and remain bounded-domain checks.

\paragraph{Baseline Prototypes.}
The Handa--Rinard rows are objective-level reproductions of published loss/complexity objectives over bounded finite weighted version spaces, supported by parity tests against direct enumeration.
They reproduce the published objectives because the public search found no original implementation.

\paragraph{LLM Scope.}
The LLM-PBE benchmark uses 20 tasks, one clean and one attacked exact-output prompt per task, deterministic answer extraction, local Ollama models, and six API rows: Qwen3.7-plus, Kimi-k2.7-code, GLM-5.2, \texttt{gpt-5.5}, \texttt{claude-opus-4-8}, and \texttt{gemini-3.1-pro-high}.
The harness is a scope demonstration for exact-output prompting, not a broad LLM benchmark.

\paragraph{Resource and Reproducibility Limits.}
Resource guards cover public SyGuS sweeps and local LLM runs; the paper-only path reruns low-memory checks and validation.

\section{Related Work}

\paragraph{PBE and Program Synthesis.}
PBE builds on version-space learning and programming-by-demonstration foundations~\cite{mitchell1982generalization,lau2003version}; FlashFill, FlashMeta, Sketch, oracle-guided synthesis, SyGuS, type/example systems, data-structure synthesis, and refinement-type systems define the main symbolic setting~\cite{gulwani2017program,gulwani2011flashfill,polozov2015flashmeta,solar2006sketch,jha2010oracle,sygus,osera2015type,feser2015datastructure,polikarpova2016refinement}.
Machine teaching, teaching dimension, active learning, RobustFill, DeepCoder, grammar/RL synthesis, blended abstract semantics, DreamCoder, and LLM-PBE connect examples to identification and learned search~\cite{goldman1995complexity,zhu2015machine,angluin1988queries,settles2009active,devlin2017robustfill,balog2017deepcoder,bunel2018grammar,nye2021blended,ellis2021dreamcoder,li2024pbe}; Raychev et al. study noisy datasets, Handa and Rinard give the closest noisy-PBE objectives and guarantees, and Rose adds abstraction refinement~\cite{raychev2016noisy,handa2020noisy,handa2021guarantees,handa2021rose}.
Raza and Gulwani give disjunctive synthesis, a robust PBE approach that partitions examples so separate branches cover heterogeneous cases~\cite{raza2018disjunctive}; VPA instead partitions into disjoint groups, synthesizes independently, and votes over semantic signatures with a vote-stability certificate.
\tool{} keeps the solver inspectable and studies worst-case fixed-set corruption.

\paragraph{Educational and Block-Based Program Feedback.}
Automated programming feedback and repair provide a complementary context in which examples, traces, and validation artifacts shape the learner-visible result. Clef generates feedback for competition-level code by learning repairs from submission histories~\cite{zhang2022clef}; PyDex repairs introductory Python assignments with LLMs~\cite{zhang2024pydex}; and a systematic study of time-limit-exceeded errors analyzes another important class of online-assignment failures~\cite{zhang2025tle}. In block-based programming, ViScratch uses Scratch code and gameplay video for automated feedback~\cite{si2025viscratch}, Stitch studies step-by-step tutoring~\cite{si2025stitch}, ScratchEval defines executable LLM-repair evaluation tasks~\cite{si2026scratcheval}, EcoScratch studies cost-aware multimodal repair~\cite{si2026ecoscratch}, Raven uses video-grounded assessment~\cite{li2026raven}, ScratchWorld evaluates executable consequences in Scratch worlds~\cite{lin2026scratchworld}, and ScratchLens defines lens-parametric equivalence for Scratch programs~\cite{scratchlens}. Other work by Zhang and collaborators applies pretrained models to merge-conflict resolution and static analysis to CI configurations and silent misconfigurations~\cite{zhang2022merge,santolucito2022ci,zhang2021configx}. These systems assume a program, trace, or assignment context and then produce or validate feedback; \tool{} studies the upstream robustness boundary of a fixed PBE example set under strategic corruption before synthesis.

\paragraph{Poisoning, Robustness, and Prompt Security.}
Adversarial examples, poisoning/backdoors, certified defenses, smoothing, and robust statistics study prediction changes or data corruption in learned systems~\cite{szegedy2014intriguing,goodfellow2015explaining,madry2018towards,biggio2012poisoning,mei2015machine,shafahi2018poison,gu2019badnets,tran2018spectral,kurita2020weight,steinhardt2017certified,levine2021deep,wang2022finite,cohen2019smoothing,rosenfeld2020label,huber1964robust,diakonikolas2016robust}.
Code-completion poisoning, TrojanPuzzle, dataset poisoning, indirect prompt injection, retrieval poisoning, in-context poisoning, and neural synthesis robustness are adjacent surfaces~\cite{schuster2021autocomplete,aghakhani2023trojanpuzzle,carlini2024poisoning,greshake2023indirect,lin2025racg,corrupt-rag,poisonedrag,he2025iclpoison,anand2021adversarial}; VPA changes partition-aggregation votes to finite PBE semantic signatures.

\section{Conclusion}

Fixed-set worst-case PBE corruption is a distinct robustness problem: strategic edits expose low-margin failures across generated DSL tasks, accepted public SyGuS slices, Playgol, noisy-PBE baselines, and LLM-PBE prompts.
VPA recovers only when target semantics keep a vote margin; on realistic public tasks with near-one margins, an adaptive attacker defeats it.

\end{document}